%% file: root.tex
\documentclass[letterpaper, 10 pt, conference]{ieeeconf}  

\IEEEoverridecommandlockouts                              

\usepackage{multicol}
\usepackage[colorlinks=true, linkcolor=blue, bookmarks=true, urlcolor=blue, anchorcolor=blue]{hyperref}
\usepackage{graphicx}
\usepackage{amsmath}
\usepackage{amssymb}
\usepackage{booktabs}
\usepackage{caption}
\usepackage{multirow}
\usepackage{color}
\usepackage{ulem}
\usepackage{float}
\usepackage{bbding}
\usepackage{multirow}
\usepackage{colortbl}
\usepackage{tikz}
\usepackage{chapterbib}    
\usepackage[noadjust]{cite}
\usepackage{kantlipsum}
\usepackage{cuted}

\usepackage[figuresright]{rotating}
\usepackage{subcaption}

\title{\LARGE \bf
RCareWorld: A Human-centric Simulation World\\ for Caregiving Robots
}

\author{Ruolin Ye$^{*1}$, Wenqiang Xu$^{*2}$, Haoyuan Fu$^{2}$, Rajat Kumar Jenamani$^{1}$, Vy Nguyen$^{1}$, \\Cewu Lu$^{2}$, Katherine Dimitropoulou$^{3}$ and Tapomayukh Bhattacharjee$^{1}$
\thanks{This work was funded by the National Science Foundation IIS (\#2132846).$^*$These two authors contribute equally to this work.}
\thanks{$^{1}$Ruolin Ye, Rajat Kumar Jenamani, Vy Nguyen, Tapomayukh Bhattacharjee are with the Department of Computer Science,
        Cornell University, Ithaca, NY, USA,
        {\tt\small \{ry273, rj277, vtn39, tb557\}@cornell.edu}}%
\thanks{$^{2}$Wenqiang Xu, Haoyuan Fu, Cewu Lu are with School of Electronic Information and Electrical Engineering, 
        Shanghai Jiao Tong University, Shanghai, China
        {\tt\small \{vinjohn, simon-fuhaoyuan, lucewu\}@sjtu.edu.cn}}%
\thanks{$^{3}$Katherine Dimitropoulou is with Columbia University, USA
        {\tt\small kd2524@cumc.columbia.edu}}%
}

\renewcommand{\baselinestretch}{0.97}

\usepackage[font=footnotesize]{caption}

\begin{document}
\maketitle




\thispagestyle{empty}
\pagestyle{empty}

\begin{abstract}
We present RCareWorld, a human-centric simulation world for physical and social robotic caregiving designed with inputs from stakeholders. 
RCareWorld has realistic human models of care recipients with mobility limitations and caregivers, home environments with multiple levels of accessibility and assistive devices, and robots commonly used for caregiving. It interfaces with various physics engines to model diverse material types necessary for simulating caregiving scenarios, and provides the capability to plan, control, and learn both human and robot control policies by integrating with state-of-the-art external planning and learning libraries, and VR devices. We propose a set of realistic caregiving tasks in RCareWorld as a benchmark for physical robotic caregiving and provide baseline control policies for them. We illustrate the high-fidelity simulation capabilities of RCareWorld by demonstrating the execution of a policy learnt in simulation for one of these tasks on a real-world setup. Additionally, we perform a real-world social robotic caregiving experiment using behaviors modeled in RCareWorld. Robotic caregiving, though potentially impactful towards enhancing the quality of life of care recipients and caregivers, is a field with many barriers to entry due to its interdisciplinary facets. RCareWorld takes the first step towards building a realistic simulation world for robotic caregiving that would enable researchers worldwide to contribute to this impactful field. Demo videos and supplementary materials can be found at: \url{https://emprise.cs.cornell.edu/rcareworld/}. 
\end{abstract}

\input{sections/intro}
\vspace{-2mm}
\input{sections/relatedwork}

\input{sections/rcareworld}

\vspace{-2mm}
\input{sections/human}

\input{sections/data_collection}
\input{sections/environment}
\input{sections/assistive_tasks}

\input{sections/experiment}
\input{sections/conclusion}


\vspace{-0.1cm}
\bibliographystyle{IEEEtran}
\bibliography{intro,references,sim,rl_algo,musculoskeletal}

\end{document}

%% file: sections/intro.tex
\section{Introduction}
\vspace{-1mm}
According to a survey in 2014~\cite{taylor2018americans}, 27.2\% of people living in the United States had a disability, and 24.2 million people aged 18 or older required assistance with activities of daily living.
Caregiving robots have the potential to provide assistance to enhance or prolong independence~\cite{broekens2009assistive} while reducing caregiving burden~\cite{chio2006caregiver}. However, building safe and meaningful robotic solutions for caregiving is challenging. The lack of access to individuals with mobility limitations and their caregivers hinders establishment of realistic design expectations. The large cost associated with real robot hardware development and maintenance is prohibitive. Simulation platforms that can realistically model care recipients, caregivers, robots, and the interactions between them in real-life caregiving scenarios can help lower these barriers to entry, and democratize this impactful field. 

\begin{figure}[!t]
\centering
\includegraphics[width=0.98\columnwidth]{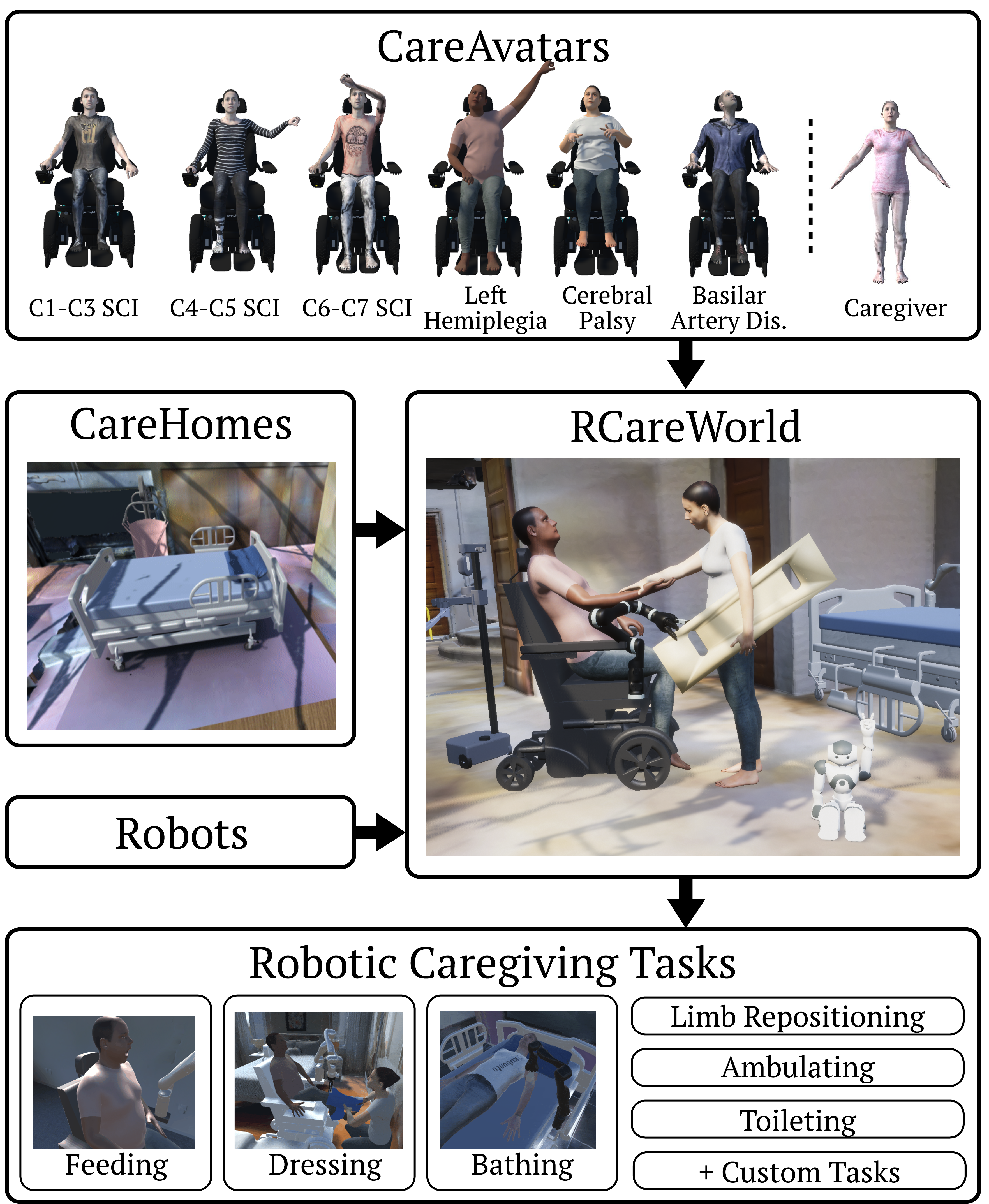}
\captionof{figure}{\small{\textbf{RCareWorld}: A human-centric simulated world for caregiving robots built using a stakeholder-informed design. 
}}
\label{fig:teaser}
\vspace{-8mm}
\end{figure}






\textit{What should an ideal simulated world for robotic caregiving have?}
By collaborating with occupational therapists, caregivers, and care recipients, we identify the necessity to (i) build realistic human avatars of care recipients and caregivers, and (ii) accurately model their home environments. These human avatars must model clinical measures of their functional abilities such as range of motion of their joints (ROM testing \cite{rom_testing}) and muscle strength (manual muscle testing \cite{mmt1}). Their home environment models must mirror the various modifications and assistive devices such as Hoyer lift, transfer board, etc. to increase accessibility.
As roboticists, we also identify the need to (iii) simulate human-robot interaction scenarios with multiple humans and robots in a scene.
While many simulation platforms have been proposed \cite{opensim, anybody, gazebo, ai2thor, habitat, sapien, virtualhome, vrkitchen, threedworld, igibson2, rfuniverse, higgins2021towards, sigverse, armsym, assistive_gym}, none address all of these needs to genuinely capture the world of robotic caregiving.

\input{tables/simulator_comparision}

In this paper, we present \textbf{RCareWorld}, a human-centric physics-based simulation world for robotic caregiving. 
Using clinical data collected from 
care recipients with mobility limitations and their caregivers, we construct \textbf{CareAvatars} -- high-fidelity models that capture the shape, movement, physics, physiological condition, and behavior of their human counterparts. We use SMPL-X models~\cite{smplx} to represent the human avatars’ shape, and integrate it with an actuable musculoskeletal or skeletal and soft-tissue backbone to model their movement and physics. We use behavior trees to model their physiological condition and behavior. 
We embed these avatars in \textbf{CareHomes} -- environment assets adapted from MatterPort3D \cite{matterport3d} to model caregiving homes with varying levels of accessibility, each containing various assistive devices. 
Our simulation world contains robots commonly used for caregiving such as Kinova Gen3, Stretch RE1, Nao, etc. It uses high-fidelity physics simulations for diverse material types, and supports various sensing modalities. Furthermore, it  supports ROS \cite{ros} and interfaces with OMPL \cite{ompl}, MoveIt \cite{moveit} for planning, Gym-style environments \cite{gym}, Stable-Baselines3 \cite{stable-baselines3} for reinforcement learning, and seamlessly integrates with VR devices. This allows for planning and control of both robot and human joints. RCareWorld is a novel simulation platform that brings together \textbf{CareAvatars}, \textbf{CareHomes}, and caregiving robots to provide a comprehensive simulation of each component in a robotic caregiving scenario.

We propose six benchmark physical robotic caregiving tasks in RCareWorld that can provide meaningful assistance to the modeled care recipients -- feeding, bathing, dressing, limb repositioning, ambulating, and toileting. We provide baseline policies for these tasks using reinforcement learning methods. To validate the fidelity of RCareWorld, we demonstrate the execution of a policy learnt in simulation for bed-bathing assistance on a real-world setup. We also perform a social robotic caregiving experiment, demonstrating the usage of behavior trees and VR integration in RCareWorld.


We summarize our contributions as follows:
\begin{itemize}
    \item \textbf{RCareWorld} -- A simulation platform that brings together human avatars (\textbf{CareAvatars}) and caregiving robots in caregiving home environments (\textbf{CareHomes}) to create a human-centric virtual caregiving world. 
    \item \textbf{CareAvatars} -- Realistic human avatars built using clinical data collected from care recipients and caregivers that capture their appearance, movement, physics, physiological condition, and behavior.
    \item \textbf{CareHomes} -- Caregiving home environment assets with varying levels of accessibility and containing various assistive devices. 
    \item We propose a set of physical robotic caregiving tasks in RCareWorld and provide baseline reinforcement learning control policies for them.
    \item We perform physical and social robotic caregiving experiments in the real world using policies learned or programmed in RCareWorld.
\end{itemize}

RCareWorld is available on our website~\cite{rcareworldweb}. We hope that making this platform accessible to researchers worldwide is a step towards democratizing the impactful field of robotic caregiving.



%% file: tables/simulator_comparision.tex
\definecolor{biomech}{HTML}{ffc09f}
\definecolor{nav}{HTML}{fcf5c7} 
\definecolor{hri}{HTML}{c0d6df}
\definecolor{ours}{HTML}{adf7b6}
\begin{table*}[ht!]
\caption{Comparison among RCareWorld and other simulation environments.}\label{tab:simulators_comparision}
{\centering

\setlength\extrarowheight{1mm}
\resizebox{\textwidth}{27mm}{
\begin{tabular}{|c|c|c|c|cccc|ccc|c|c|c|c|c|} 
\hline
\multirow{2}{*}{Name} & \multirow{2}{*}{\begin{tabular}[c]{@{}c@{}}Env. \\Asset$^1$\end{tabular}} & \multirow{2}{*}{\begin{tabular}[c]{@{}c@{}}Assistive\\Devices\end{tabular}} & \multirow{2}{*}{\begin{tabular}[c]{@{}c@{}}Human \\Model$^2$\end{tabular}} & \multicolumn{4}{c|}{Human Movement} & \multicolumn{3}{c|}{Robot Movement} & \multirow{2}{*}{Sensing $^5$} & \multirow{2}{*}{ROS} & \multirow{2}{*}{Object Types$^6$} & \multirow{2}{*}{\begin{tabular}[c]{@{}c@{}}Physics\\Backends$^7$\end{tabular}} &
\multirow{2}{*}{\begin{tabular}[c]{@{}c@{}}Offscreen \\Rendering\end{tabular}} \\ 
\cline{5-11}
 &  &  &  & Planning & Learning & VR$^3$ & Animation$^4$ & Planning & Learning & VR &  &  &  &  &  \\ 
\hline
\rowcolor{biomech} OpenSim \cite{opensim} & - & - & M & - & - & - & \checkmark & \multicolumn{3}{c|}{-} & RGB,~JF/T* & - & - & Simbody & - \\
\rowcolor{biomech} AnyBody \cite{anybody} & - & - & M & - & - & - & \checkmark & \multicolumn{3}{c|}{-} & RGB,~JF/T & - & - & - & - \\ 
\hline
\rowcolor{nav} Gazebo \cite{gazebo} & - & - & S & - & - & - & \checkmark & \checkmark & \checkmark & - & RGBD,~JF/T* & \checkmark & R, A, S & ODE, B, D, Simbody & \checkmark \\
\rowcolor{nav} AI2THOR series \cite{ai2thor} & \checkmark* & - & - & \multicolumn{4}{c|}{-} & - & \checkmark & - & RGBD,~JF/T* & - & R, A & P & \checkmark \\
\rowcolor{nav} Habitat \cite{habitat, chen2021ros} & \checkmark* & - & - & \multicolumn{4}{c|}{-} & \checkmark* & \checkmark & - & RGBD, JF/T* & \checkmark* & R, A & B & \checkmark \\
\rowcolor{nav} SAPIEN \cite{sapien} & - & - & - & \multicolumn{4}{c|}{-} & \checkmark & \checkmark & - & RGBDN,~JF/T* & \checkmark & R, A & P & \checkmark \\
\rowcolor{nav} VirtualHome \cite{virtualhome} & \checkmark* & - & S & - & - & - & \checkmark & \multicolumn{3}{c|}{-} & RGBN,~JF/T* & - & R, A, D & P & - \\
\rowcolor{nav} VRKitchen \cite{vrkitchen} & \checkmark** & - & S & - & - & \checkmark & - & \multicolumn{3}{c|}{-} & RGBD,~JF/T* & - & R, A, D, F & P, & - \\
\rowcolor{nav} ThreeDWorld \cite{threedworld} & \checkmark* & - & - & \multicolumn{4}{c|}{-} & - & \checkmark & \checkmark & RGBD,~JF/T* & - & R, A, D, F & P,~F & - \\
\rowcolor{nav} Gibson series \cite{igibson2}& \checkmark* & - & - & \multicolumn{4}{c|}{-} & \checkmark & \checkmark & - & RGBD,~JF/T* & - & R, A & B & \checkmark \\
\rowcolor{nav} RFUniverse \cite{rfuniverse}& - & - & - & \multicolumn{4}{c|}{-} & \checkmark & \checkmark & \checkmark & RGBDN, JF/T & - & R, A, F, D, Te & P, O, F, CD & \checkmark \\ 
\hline
\rowcolor{hri} RIVR \cite{higgins2021towards} & - & - & S & - & - & \checkmark & - & \checkmark & \checkmark & \checkmark & RGBD, JF/T* & \checkmark & R, A & P & - \\
\rowcolor{hri} SIGVerse \cite{sigverse} & \checkmark* & - & S & - & \checkmark & \checkmark & \checkmark & \checkmark & \checkmark & \checkmark & RGBD, JF/T* & \checkmark & R, A & P & \checkmark \\
\rowcolor{hri} ArmSym \cite{armsym} & - & - & S & - & - & \checkmark & - & - & - & \checkmark & RGB & - & R & P & - \\
\rowcolor{hri} Assistive Gym \cite{assistive_gym} & - & - & S & - & \checkmark & \checkmark* & - & - & \checkmark & \checkmark$\dagger$ & RGBD, JF/T & - & R, A, D & B & \checkmark \\ 
\hline
\rowcolor{ours} RCareWorld & \checkmark & \checkmark & M & \checkmark & \checkmark & \checkmark & \checkmark & \checkmark & \checkmark & \checkmark & RGBDN, JF/T, DF & \checkmark & R, A, F, D, Te & P, O, F, CD, Z, S & \checkmark \\
\hline
\end{tabular}}}
\\[3pt]

\tikz\draw[biomech,fill=biomech] (0,0) circle (.8ex); for biomechanical simulators for human modeling, \tikz\draw[nav,fill=nav] (0,0) circle (.8ex); for simulation environments for navigation and manipulation, \tikz\draw[hri,fill=hri] (0,0) circle (.8ex); for simulation environments for human-robot interaction, and \tikz\draw[hri,fill=ours] (0,0) circle (.8ex); for RCareWorld.\\
We only 
\textbf{1}: \checkmark* means the simulation environments provide household environment assets, while \checkmark means the simulation environments provide realistic caregiving environments with home modification.\\
\textbf{2}: M for Musculoskeletal, and S for Skeletal.\\
\textbf{3}: \checkmark$\dagger$ in this column means the feature is added in the following work.\\
\textbf{4}: Animation includes MoCap data replaying and IK-based motion generating.\\
\textbf{5}: RGB for RGB image, \textbf{D} for \textbf{D}epth, \textbf{N} for Surface \textbf{N}ormal and \textbf{JF/T} for \textbf{J}oint \textbf{F}orce/\textbf{T}orque sensors. \textbf{JF/T*} means force/torque sensors that are not explicitly implemented in the simulation environment, but can be easily obtained with corresponding physics backends. \textbf{DF} for \textbf{D}istributed \textbf{F}orce sensing.\\
\textbf{6}: \textbf{R} for \textbf{R}igid, \textbf{A} for \textbf{A}rticulated, \textbf{D} for \textbf{D}eformable, \textbf{F} for \textbf{F}luid, and \textbf{Te} for \textbf{Te}arable objects. \textbf{Te*} means the manipulation operation, eg. slicing, for tearable objects in this simulation environment is animated rather than physics-based.\\
\textbf{7}: \textbf{B} for \textbf{B}ullet, \textbf{P} for \textbf{P}hysX, \textbf{F} for \textbf{F}lex, \textbf{O} for \textbf{O}bi, \textbf{CD} for \textbf{C}loth \textbf{D}ynamics, \textbf{Z} for \textbf{Z}ibra Liquids, and \textbf{S} for \textbf{S}OFA, \textbf{ODE} for \textbf{ODE}.
\vspace{-6mm}
\end{table*}

%% file: sections/relatedwork.tex
\section{Background}
\label{sec:related_work}
\vspace{-1mm}
Realistic modeling of humans and robots is paramount for simulating physical and social robotic caregiving scenarios. While existing biomechanical simulators tackle high-fidelity human modeling (Sec.~\ref{human_model}), and robotics simulators provide realistic simulation of robots (Sec.~\ref{simulation_envs}), there is hardly any work at the intersection that captures both of these worlds well enough for our utility.

\subsection{Human Modeling}
\label{human_model}
Realistic human modeling comprises modeling shape, movement, physics, physiological condition, and behavior.

\textbf{Shape.} 
Shape models for the human body capture dimensions of major bones of the body along with the 3D surface of skin \cite{joo2018total,smplx}.
We adopt the SMPL-X model \cite{smplx} as it has a simple parametrization, and several methods exist \cite{rong2021frankmocap} to regress its parameters from RGB images or RGB-D data. 


\textbf{Movement and Physics.} 
Movement in human body mostly happens due to the musculoskeletal system. Musculoskeletal physics can be modeled by muscle-actuated joint dynamics. Many biomechanical simulators \cite{anybody,opensim} adopt a hill-type muscle model \cite{hilltype0} because of its balance between computational complexity and accuracy \cite{hilltype1,hilltype3}. We use hill-type muscles to generate muscle-actuated movement with data from \cite{opensim}, and simulation framework from \cite{hilltype2}.

Modeling soft tissues are an alternative way to model the physics in musculoskeletal system. Soft-body deformation and forces can be computed by various methods such as XPBD \cite{xpbd}, FEM \cite{artisynth}, or neural network-based representations \cite{kim2017data}. We consider soft tissues in the human body such as skin, muscle, and fat as soft bodies. In practice, since FEM-based approaches are rather slow, while neural network-based approaches require training, we adopt an XPBD-based solver \cite{obi}.

\textbf{Physiological Condition and Behavior.} 
Hester et al.~\cite{hum_mod} model and simulate human physiological conditions, and describe them as variables.
In this paper, we primarily consider gestures as a medium to express human behavior. Gestures can be modeled as an organized sequence of physical motions. It can be described by models such as finite state machines or behavior trees \cite{iovino2020survey}. We adopt behavior trees since they provide a simple way to model complex behaviors including handling parallel tasks and reactions. 

\subsection{Simulation Environments}
\label{simulation_envs}
Simulation environments are important tools for almost every sub-field of robotics research. We show a comprehensive comparison in Table \ref{tab:simulators_comparision}.

\textbf{Navigation and Manipulation.} With a recent explosion of interest in robot learning, many simulation environments have been proposed to facilitate research on navigation and manipulation tasks \cite{gazebo, ai2thor, habitat, sapien, virtualhome, vrkitchen, threedworld, igibson2, rfuniverse}. They usually adopt off-the-shelf physics backends such as ODE \cite{ode}, PhysX \cite{PhysX}, and Bullet \cite{pybullet} for physics simulation, high-quality renderers for visual perception, and provide model assets for robots, objects, and homes. 
RCareWorld is Unity-based, and extends the Python-Unity communication interface of RFUniverse \cite{rfuniverse} which is built upon ML-Agents \cite{mlagents}.

\textbf{Human-Robot Interaction.}
Simulation environments for human-robot interaction typically use virtual reality (VR) interfaces to control virtual human avatars 
\cite{armsym,higgins2021towards,sigverse}
. The data produced during interaction can be used as demonstrations to potentially guide policy learning for human-robot collaboration tasks in the real world \cite{vrkitchen, watch_and_help}. For assistive human-robot interaction, Assistive Gym \cite{assistive_gym} provides Gym-style environments for reinforcement learning. It represents human avatars through simplified shape primitives that are assumed to be rigid articulated objects. However, RCareWorld provides support for high-fidelity human avatars with realistic musculoskeletal systems and soft-tissue modeling. It also has realistic home environments with different levels of modifications, and assistive devices.

%% file: sections/rcareworld.tex
\input{fig/system_overview}
\section{RCareWorld: A Human-centric Simulation World for Caregiving Robots}\label{sec:rcareworld}

RCareWorld is a novel simulation platform that brings \textit{human avatars} and \textit{robots} together in \textit{assistive environments} to create a simulation world for physical and social caregiving. For human avatars, we propose \textbf{CareAvatars} that realistically model the shape, movement, physics, physiological condition, and behavior of their human counterparts (Sec. \ref{sec:human}). For robots, RCareWorld supports robots that are commonly used in physical and social caregiving settings. Its supports Kinova Gen3, Kinova Gen2, Stretch RE1, Franka Emika Panda, UR5, PR2, Pepper robot, HSR, Fetch, and NAO robot. Users can add new robots of their choice by using URDF files and setting movable joints. For assistive environments, we propose \textbf{CareHomes} with home modifications at multiple levels of accessibility including various assistive devices (Sec. \ref{sec:human}). Figure ~\ref{fig:components} shows the overview of system components of RCareWorld. It uses Unity as a server, and Python, ROS, and VR as clients. 

RCareWorld's Unity server is responsible for high-fidelity physics simulation, photorealistic visual rendering, and multimodal sensing. It interfaces with various physics engines to model diverse material types necessary for simulating caregiving scenarios such as rigid, articulated, fluid, deformable, and tearable materials. It uses Unity's Scriptable Rendering Pipeline for visual rendering, and supports RGB, depth, surface normal, joint force/torque, and contact force sensing. Agents can be controlled through their joints' positions / velocities / torques and muscle activation (for musculoskeletal \textbf{CareAvatars}). 

The Unity server communicates with the Python client using gRPC, the ROS client with Unity Robotics Hub, and the VR devices with SteamVR. Users can access state and sensory information, and control human and robot agents using these interfaces. Python provides unique support for learning tasks by providing a Gym wrapper which enables users to leverage reinforcement learning libraries such as Stable-Baslines3~\cite{stable-baselines3}. ROS makes it easier for users to perform motion planning with existing libraries including MoveIt!~\cite{moveit} and OMPL~\cite{ompl}. The VR interface aims to bridge the gap between real world and simulated world. It enables users to interact with the simulated world, and is conducive for research in social robotic caregiving. It also makes it possible to collect demonstration data to learn from.

%% file: fig/system_overview.tex
\begin{figure}[!t]
\begin{center}
\includegraphics[width=0.95\linewidth]{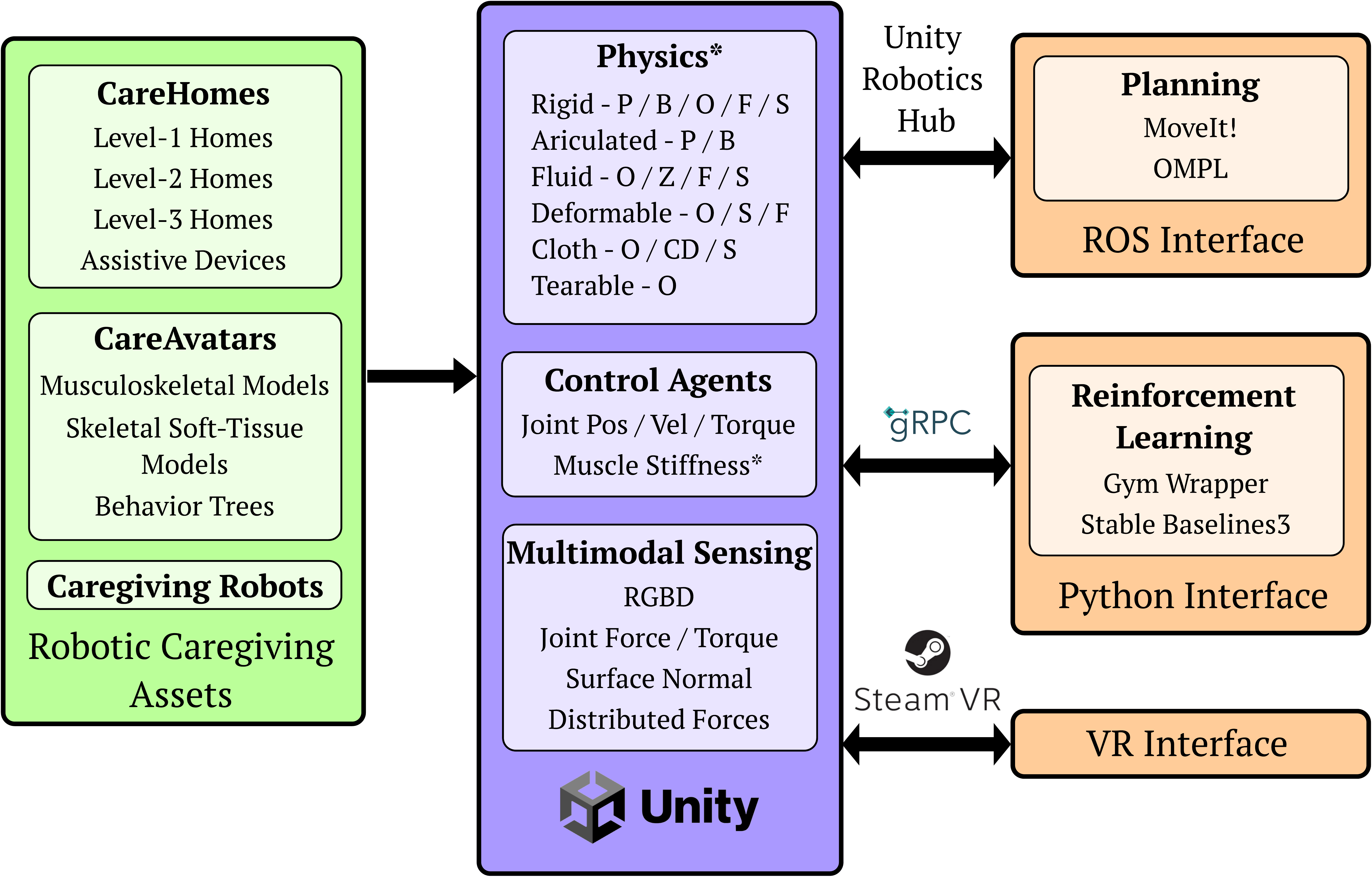}
\end{center}
\vspace{-2mm}
\centering
*\scriptsize{\textbf{P}hysX, \textbf{B}ullet, \textbf{O}bi, \textbf{F}lex, \textbf{C}loth \textbf{D}ynamics, \textbf{Z}ibra Liquids, and \textbf{S}OFA}
\vspace{-1mm}
\caption{System overview of RCareWorld.
}
\label{fig:components}
\vspace{-8mm}
\end{figure}

%% file: sections/human.tex
\section{CareAvatars: Human Modeling}\label{sec:human}
Realistic human models need to comprehensively model the shape and structure, movement, physics, physiological condition, and behavior of the human body. We propose a novel musculoskeletal SMPL-X model with behavior trees for modeling these aspects. We then use these models to build virtual avatars for one caregiver and six care recipients having varying disabilities and caregiving needs.

\begin{figure}[!t]
\centering
\includegraphics[width=\columnwidth]{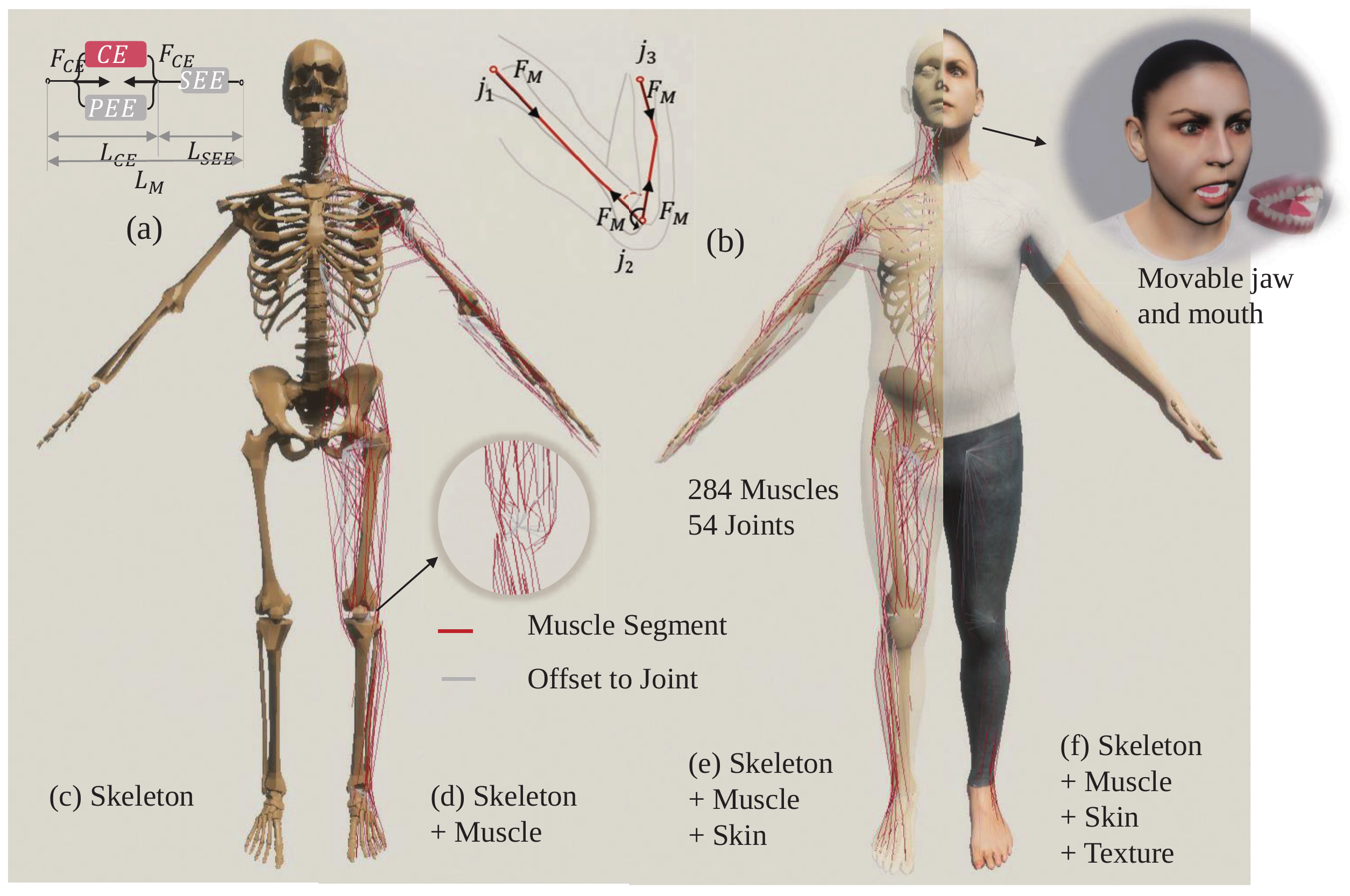}
\captionof{figure}{\footnotesize{\textbf{CareAvatars}: Visualization of musculoskeletal SMPL-X models. (a) Muscle segment: A muscle segment consists of three elements CE, PEE, and SEE, (b) Muscle force $F_M$ actuates the skeletal system, (c) Skeletal human model, (d) Skeletal human model with muscles, (e) Skeletal human model with muscles and skin, (f) Skeletal human model with muscles, skin, and texture.}}
\label{fig:teaser}
\vspace{-7mm}
\end{figure}

\subsection{Realistic Human Models}
By `realistic', we mean realism in shape structure and physics. For visual realism, we use textures from \cite{surreal, smplx}.
\subsubsection{Model Structure}\label{sec:smplx}
\begin{figure}[!t]
\begin{center}
\includegraphics[width=\linewidth]{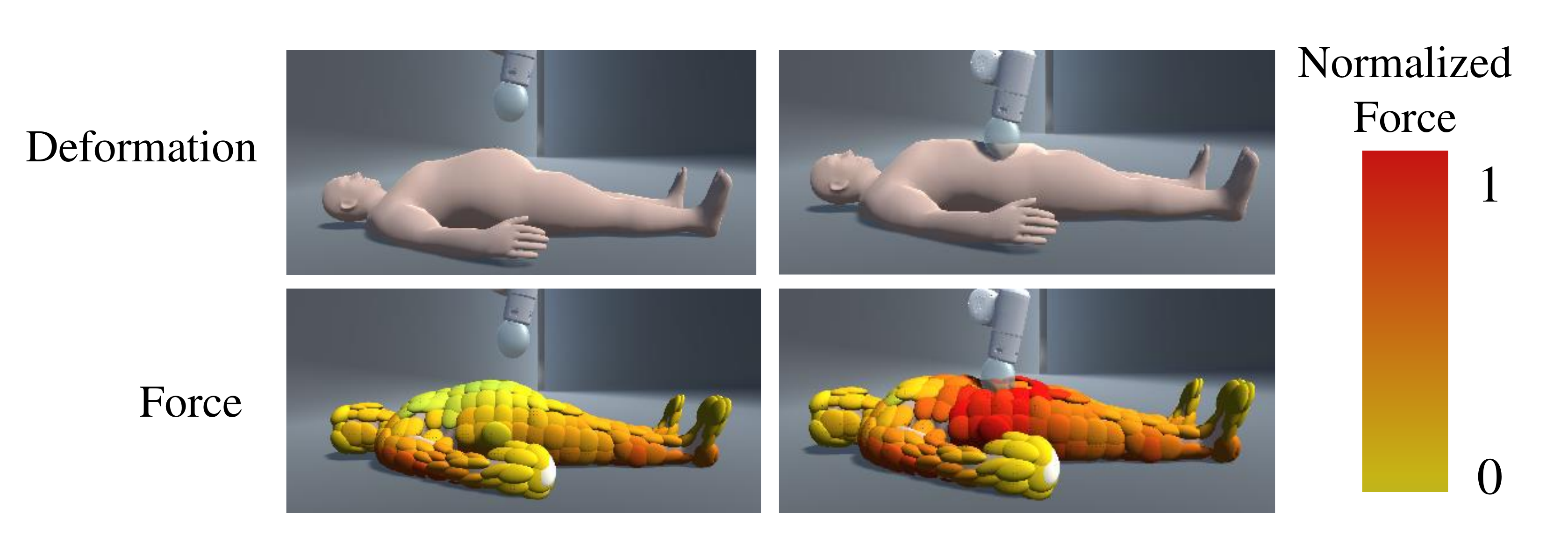}
\end{center}
\vspace{-5mm}
\caption{\footnotesize{\textbf{CareAvatars}: Soft-tissue deformation and force due to external contact. The top row shows deformation of human skin, and the second row shows force modeled with XPBD. We normalize the forces to be in the range from 0 to 1. }}
\label{fig:pressure}
\vspace{-8mm}
\end{figure}

We model two types of actuation methods, one that uses musculoskeletal actuation, and the other that uses skeletal actuation with soft tissues. The musculoskeletal actuation system (Fig. \ref{fig:teaser}) uses muscles on top of a skeleton and a skin for actuation, while the skeletal actuation system uses skeleton movements to actuate the skin and soft tissues (Fig. \ref{fig:pressure}).

\paragraph{Skeleton and Skin}
Both our human model actuation methods use skeleton composed of bones and joints as well as skin. Following the SMPL-X model~\cite{smplx}, we define a human mesh model as $M(\theta, \beta, \psi): \mathbb{R}^{|\theta| \times|\beta| \times|\psi|} \rightarrow \mathbb{R}^{3 N}$, where $N=10475$ is the number of vertices in the 3D mesh of a body. $\theta \in \mathbb{R}^{3(K+1)}$ is the parameter for human pose, where $K=\theta_f + \theta_h + \theta_b=54$ is the number of body skeleton joints, and there is an additional joint for global rotation. $\theta_f=3$ 
are the joints for jaw and eyeballs, $\theta_h=30$ are the joints for the fingers, and $\theta_b=21$ are the joints for the remaining body skeleton. Each joint is modeled with a universal joint with 3 degrees of freedom. $\beta \in \mathbb{R}^{|\beta|}$ is the parameter for body shape, and $\psi \in \mathbb{R}^{|\psi|}$ is for facial expression.

\paragraph{Musculoskeletal Actuation}
The actuation system of muscle on joints is modeled by a typical 3-element Hill-type model \cite{hilltype0}. As shown in Fig. \ref{fig:teaser}, 
a muscle is linked by muscle segments, and the muscle segment is visualized in red line, and modeled by three elements, namely the contractile element $CE$, the parallel elastic element $PEE$, and the serial elastic element $SEE$. To construct a whole-body musculoskeletal system for SMPL-X, we refer to the biological structure of muscles (i.e. the origin, pathway, insertion points) \cite{stewart2009skeletal}, and the data from MASS \cite{hilltype3}. Furthermore, we convert conventional bone-based muscle positioning to joint-based muscle positioning, which means that each muscle segment is located at an offset from a specific joint. In this way, we can adapt muscles according to joint variations.

\paragraph{Skeletal Actuation with Soft Tissues}
For the skeletal actuation system that uses soft tissues, we assume the soft tissue attached to the skeleton system to be isotopic in our human models. We represent them by a volume mesh, and model them with Extended Position-Based Dynamics (XPBD) \cite{xpbd}. Fig. \ref{fig:pressure} shows the deformation of soft tissues when a robot touches a human's arm with its end-effector. 

\subsubsection{Movement Modeling - Kinematics}
\label{sec:kinematics}

\paragraph{Joint Limits}
For the human musculoskeletal system, different levels and categories of mobility limitations can affect a person's range of motion (ROM).
Clinical research categorizes ROM into Active ROM (AROM) and Passive ROM (PROM).
AROM is the range of motion that a care recipient can achieve by themselves for a specific joint. PROM is the range of motion of a joint when an external caregiver produces the movement. Care recipients with different ROM need different care strategies \cite{rom6}. We set PROM and AROM in the URDF files based on clinical data.

\paragraph{Inverse Kinematics and Full-Body Motion Planning}
We adopt BioIK \cite{bioik} for human full-body inverse kinematics calculation. 
It can be well integrated into both MoveIt! \cite{moveit} for motion planning, and Unity for IK-based animation.


\subsubsection{Physics Modeling - Dynamics}
\paragraph{Inverse Dynamics for Bones}
When human avatars in simulation are actuated by pure skeletons, the dynamics is usually simpler compared to actuation using muscles, but the resulting actuation may not be as realistic.
For each joint $k$ in skeletal actuation, we can describe its dynamics by 
$
M(q_k) \ddot{q_k}+C(q_k, \dot{q_k})q+G(q_k)=\tau_k
$ where $M(q)$ is mass, $C(q, \dot{q})$ is Coriolis term, and $G(q)$ is the gravitational term. $\tau_k$ is the applied torque at joint $k$ and $q_k$, $\dot{q_k}$, and $\ddot{q_k}$ are the joint position, velocity, and acceleration respectively. 

\begin{figure*}[!t]
\begin{center}
\includegraphics[width=0.95\linewidth]{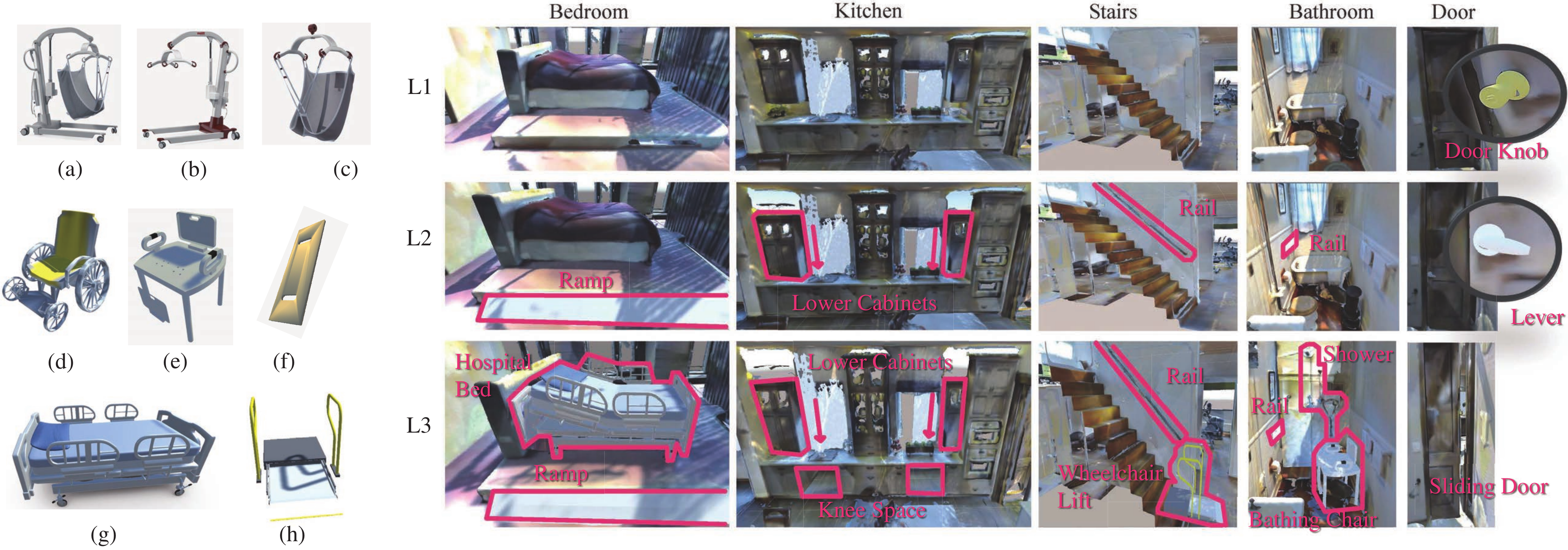}
\end{center}
\vspace{-3mm}
    \caption{\footnotesize\textbf{CareHomes:} \textit{Left:} Assistive devices supported in RCareWorld (a) Hoyer lift with a universal sling, (b) Hoyer lift for transfers, (c) Ceiling lift for a sling, (d) Manual wheelchair, (e) Transfer bench bathtub chair, (f) Transfer slide board, (g) Hospital bed for home, (h) Wheelchair lift with ramp.  \textit{Right:} L1: 'Usual` - home assets with no specific modifications, L2: 'Partially barrier-free' - some frequently used objects have modifications, L3: 'Completely barrier-free' - all possible modifications are utilized. We model these using Universal Design Manual~\cite{universal_design} and inputs from occupational therapists.}
\label{home_modification}
\vspace{-5mm}
\end{figure*}

\paragraph{Inverse Dynamics for Muscles}
For a musculoskeletal actuation system using a Hill-type model, we can calculate joint torques with respect to muscle activation signals. Using the three elements ($CE$, $PEE$, and $SEE$) shown in Fig. \ref{fig:teaser}, we can calculate the total contractile force $F_M$ as a function of the contractile force $F_{CE}$ and muscle states $x$ ~\cite{muscle_biped} given by 
$
 F_{\mathrm{M}} = f(F_{\mathrm{CE}}, x)
$ 
 ,where $F_{CE}$ is given by
\begin{equation}\label{eq:3}
F_{\mathrm{CE}}=a \mathrm{~F}_{\max } f_{\mathrm{L}}\left(L_{\mathrm{CE}}\right) f_{\mathrm{V}}\left(V_{\mathrm{CE}}\right).
\end{equation}

Here, $\mathrm{~F}_{\max }$ is the constant maximum isometric force for the muscle, $f_{\mathrm{L}}$ describes the relationship between force and length of a muscle, and $f_{\mathrm{V}}$ describes the relationship between force and the current contraction velocity $V_{CE}$. $L_{CE}$ is the length of $CE$, and $a$ is the muscle activation. 
For each joint $k$, a torque $\tau_{k}$ is generated by
$
\tau_{k}=F_{M}\left\|\left(p-j_{k}\right) \times \frac{\mathbf{s}_{c}}{\left\|\mathbf{s}_{c}\right\|}\right\| 
$ 
where $p$ is the point where the muscle is attached to the bone, $j_k$ is the joint to apply the torque on, and $s_c$ is the muscle segment. 


\subsubsection{Physiological Condition and Behavior Modeling}
Physiological condition is one of the important representations of human well-being. We provide an interface to set body temperature, blood pressure, and heart rate for human avatars, which are variables that can be defined by a user, and modified during simulation. For human behavior modeling, we take advantage of the composability and modularity of behavior trees. With hierarchical task nodes and control flow, sub-trees for a simple task can be composed into more comprehensive and complex behaviors. Also, a sub-tree for a task can be reused for different agents. 







%% file: sections/data_collection.tex
\subsection{Building Virtual Avatars}

We use our proposed human models to build realistic virtual avatars for one caregiver and six care recipients having varying causes of disabilities and caregiving needs. We call these avatars \textbf{CareAvatars}, and use publicly available clinical data from SPARCS-box~\cite{sparcs} (Table \ref{tab:six_avatar}) to create them. For each of these care recipients, SPARCS-box provides body dimensions, weight, AROM, PROM, and MMT data. We describe how we map this clinical data below: 

\subsubsection{Skeleton and Skin}
We generate the skeletal parameters for our human models for each care recipient from their body dimensions, and randomly select their shape parameters. The muscle connections within the skeleton for our musculoskeletal SMPL-X models are self-adaptive. 

\subsubsection{ROM}
We ignore the eyeball joints, and set ROM values for the other 52 joints in our human models by generating two URDF files for each avatar, one each for AROM and PROM.

\subsubsection{MMT}
We adopt a uniform linear mapping to map the MMT scale to constant maximum isometric force $F_{max}$ as shown in Eq. \ref{eq:3}. We map MMT grades to scaled average $F_{max}$ mentioned in \cite{hilltype2}.

\subsubsection{Mass}
Following anthropomorphic data (details in our website~\cite{rcareworldweb}), we set the weight of different body parts according to their percentage of the whole-body weight.

\begin{table}[t]
\centering
\vspace{-0.1cm}
\caption{Information for the six care recipients~\cite{sparcs}}
\vspace{-0.1cm}
\label{tab:six_avatar}
\begin{tabular}{cc}
\hline
Identifier & Cause of Disability                  \\ \hline
Morgan (he/him)   & Brainstem Stroke  \\
Jose (they/them)   & Spinal Cord Injury (C1-C3)                             \\
Natalia (she/her) & Spinal Cord Injury (C4-C5)                                \\
Daniel (he/him)   & Spinal Cord Injury (C6-C7)                             \\
Kim (she/her) & Cerebral Palsy                                   \\
Karan (he/him)   & Left-side Hemiplegia                            \\ \hline
\end{tabular}
\vspace{-0.6cm} 
\end{table}

%% file: sections/environment.tex
\begin{figure*}[!t]
\begin{center}
\includegraphics[width=0.9\linewidth]{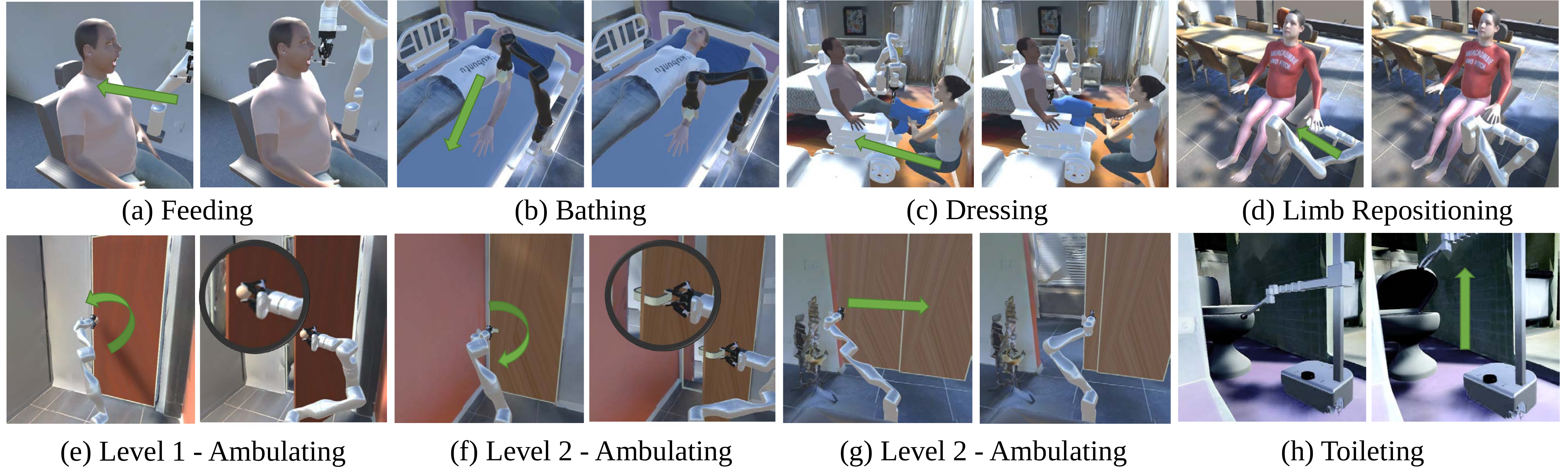}
\end{center}
\vspace{-4mm}
\caption{\footnotesize{Image sequences from executing trained robot policies. For each task, we show the initial state (\textit{left}) and the final state on task completion (\textit{right}).}}
\label{fig:rl}
\vspace{-7.5mm}
\end{figure*}

\section{CareHomes: Environment Modeling}\label{sec:env}
By consulting stakeholders, including occupational therapists, caregivers, and care recipients, and following the guidance of Universal Design Manual \cite{universal_design}, we present modifications to home environments to better align with assistive care scenarios. Some examples of home modifications are as follows: wheelchair users need wide enough hallways and doorways to navigate in and out of their living spaces. They need stairs equipped with lifts to access different floors of their home. They can benefit from knee spaces under counters and tables, and use these surfaces. In RCareWorld, we present three levels of modifications: Level 1 is ``usual'', which means no special modifications. Level 2 is ``partially barrier-free'', where we modify some objects that are frequently used. Level 3 is ``completely barrier-free''. 

We adopt the assets from Matterport3D \cite{matterport3d}, and propose a new scene asset \textbf{CareHomes}. It consists of 16 houses, including 17 kitchens, 59 bedrooms, 70 bathrooms, 17 living rooms, 16 dining rooms, 18 lounges, and 41 other kinds of rooms. For each room, we provide its corresponding level 2, and level 3 modifications in the \textbf{CareHomes} assets. Typical examples of a room in different levels are shown in Fig. \ref{home_modification}. We have also included assistive devices that are commonly used by care recipients for functioning and participation in their routines. We show the devices in Fig. \ref{home_modification}.

%% file: sections/assistive_tasks.tex
\section{Benchmarking Robotic Caregiving Tasks}\label{assistive_tasks}
We propose six physical robotic caregiving tasks in RCareWorld as a benchmark for researchers to compare reinforcement learning policies, or test planning and control algorithms. These tasks have been designed with inputs from professional occupational therapists to ensure that they can provide meaningful assistance.
For feeding, dressing, limb repositioning, and ambulating, we use a Kinova Gen3 mounted on a wheelchair. For bed bathing, we use a Kinova Gen2 grasping a sponge. For toileting, we use a Stretch RE1 robot. Videos are available on our website~\cite{rcareworldweb}.\\
\textbf{- Feeding:} The robot holds a spoon with a food-item to feed Karan, the care recipient with Left-side Hemiplegia (Table~\ref{tab:six_avatar}). The task is successful if the robot reaches the target near Karan's mouth without dropping the food. 
\\
\textbf{- Bathing:} Natalia, with C4-C5 spinal cord injury (Table~\ref{tab:six_avatar}), lies in a hospital bed. The robot holds a soft sponge to wipe her upper arm. The task is successful if the robot reaches her elbow without exceeding the safety force threshold.
\\
\textbf{- Dressing:} A caregiver lifts one leg of Daniel, with C6-C7 spinal cord injury (Table~\ref{tab:six_avatar}). The task is successful if the robot pulls the short hanging on his leg along the central axis of his leg towards his hip.
\\
\textbf{- Limb Repositioning:} Jose, with C1-C3 spinal cord injury (Table~\ref{tab:six_avatar}), sits in a wheelchair with their left arm hanging outside the armrest. This usually is the case after a transfer to the wheelchair using a hoyer sling. The task is successful if the robot lifts their arm and places it on the wheelchair handle.
\\
\textbf{- Ambulating:} Homes at different levels of accessibility in \textbf{CareHomes} have different types of doors. Level 1 homes have doors with knobs, Level 2 homes have doors with levers, and Level 3 homes have sliding doors. The task is successful if the door knob and lever rotates larger then a threshold angle, and the sliding door reaches the open position.
\\
\textbf{- Toileting:} A Stretch RE1 in a bathroom has to lift up a toilet lid. The task is successful if the lid opens larger than a threshold angle.

%% file: sections/experiment.tex
\section{Experiments and Demonstrations}
RCareWorld provides realistic human models and environments for robotic caregiving. We analyze whether these high-fidelity models affect simulation speed in Sec. \ref{speed}, benchmark the reinforcement learning pipeline in Sec.\ref{learning}, and extend the trained policies in simulation to real world in Sec.~\ref{real_robot}. Additionally, we evaluate the performance of policies programmed in RCareWorld for real-world robotic social interaction in Sec.~\ref{social}.
\subsection{Human Simulation: Trade-off between Fidelity and Speed}\label{speed}
We show the simulation speed of skeletal and musculoskeletal actuation systems where we actuated the neck pitch of one human avatar in an empty scene over 5 seconds, and computed the average frame rate. We test the speed on a Windows 10 system with an Intel i7-8750H CPU. The FPS of skeletal human simulation is 507, while that of musculoskeletal human simulation is 447. After we add soft tissues to the skeletal system, the FPS becomes 87. The result suggests our simulation for muscle-actuated joint dynamics adds little overhead to the original skeletal system, while the simulation of soft tissues has a slower speed. However, this simulation speed is sufficient for real-time performance, thus making it satisfactory for general use cases.

\subsection{Experiments using Reinforcement Learning in Simulation}\label{learning}

We provide baseline reinforcement learning policies for the six benchmark physical robotic caregiving tasks proposed in Sec. \ref{assistive_tasks}. We train these policies using implementations from Stable-Baselines3~\cite{stable-baselines3}. Table~\ref{tab:rl} shows the algorithm, reward, and success rate for each task. Some tasks have a distance to target ($d2t$) term in their reward function. We randomize these targets for robots to reach on (1) elbow surfaces for bathing, (2)  the knee for dressing, (3) the handle of the wheelchair for limb repositioning.
For each task, we train 5 policies with different random seeds. We run 100 trials with each model and report average success rates.
Figure \ref{fig:rl} demonstrates execution of the baseline policy for each task. Demonstration videos and further details of the experimental setup can be found on our website~\cite{rcareworldweb}.

\begin{table}[]
\vspace{-0.1cm}
\caption{Reinforcement Learning Simulation Results}
\label{tab:rl}
\vspace{-0.2cm}
\begin{tabular}{llp{0.4\linewidth}l}
\hline
Task       & Algorithm & Reward        & Result \\ \hline
Feeding    & SAC~\cite{sac}   & $is\_true(food\_on\_spoon)\times(is\_true(food\_on\_spoon)+is\_true(spoon\_near\_mouth))$ &  92\%\\ \hline
Bathing    & SAC~\cite{sac}   &  $-d2t - force_{end\_effector}$            &     99\% \\ \hline
Dressing   & TQC~\cite{tqc}   &    $-d2t$           &   93\%   \\ \hline
Limb Repo. & SAC~\cite{sac}   &    $-d2t$           &    65\%  \\ \hline
Ambulating & SAC~\cite{sac}   &   $knob/lever\_rotation\_angle$, or $-d2t$ &      \\ \hline
Toileting  & SAC~\cite{sac}   &    $open\_angle$           &   83\%   \\ \hline
\end{tabular}
\vspace{-0.2cm}
\end{table}

\subsection{Real-world Physical Caregiving: Sim-to-Real Transfer}\label{real_robot}

To showcase the fidelity of RCareWorld, we demonstrate the successful sim-to-real transfer of a policy learnt in this simulation platform.
\
We consider the setting for \textbf{Bathing} Natalia as proposed in Sec. \ref{assistive_tasks}. We use the same training scheme as proposed for the baseline, but update the comfortable contact force range to match the preferences of the real-world user. The user is comfortable with a contact force in range of $0-20$N. This policy in simulation has access to the ground truth pose of the robot as an observation. 

\begin{figure}[t!]
\begin{center}
\includegraphics[width=\linewidth]{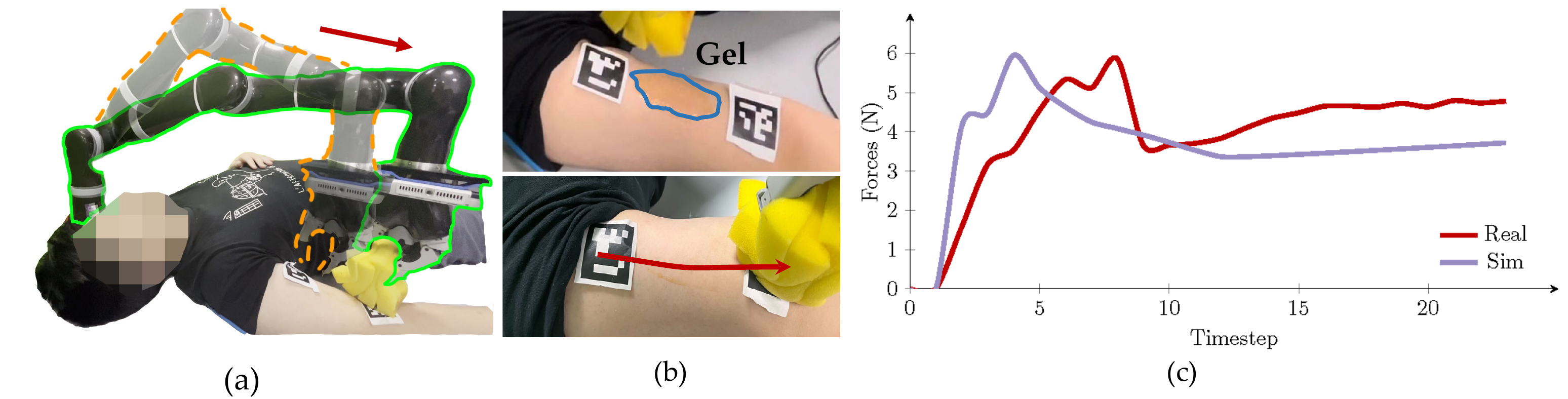}
\end{center}
\vspace{-0.4cm}
\caption{\footnotesize{Real-world Physical Caregiving: (a) the trajectory of Kinova Gen2 when executing bathing policy, (b) human arm before and after bed bathing, (c) the forces on the end-effector in simulation and in the real world.}}
\label{fig:real_robot}
\vspace{-0.8cm}
\end{figure}

We create a setup similar to the bathing environment in the real world. We identify a user whose body dimensions are similar to Natalia's. This user lies on a bed with a 7-DoF Kinova Gen2 robot mounted nearby. We additionally mount a Realsense D415 camera on the robot's wrist. We put AruCo markers on the user's left shoulder and elbow to locate the start and target position, and put colored gel on their skin for the robot to wipe off. A robot trial starts with the robot initially near the left shoulder of the user. The trial is considered a success if the robot reaches the elbow joint position, and it does not apply a force that is outside the comfortable threshold.  We perform 5 trials with the user, and note the success rate to be $80$\%. Failures due to failed pose estimation of AruCo markers have not been considered. Figure~\ref{fig:real_robot} shows our experiment setup, the trajectory executed by the robot for bed bathing, the arm before and after wiping, as well as the forces applied by the robot in simulation and in the real world.


\subsection{Real-world Robotic Social Caregiving}\label{social}
\begin{figure}[!t]
\begin{center}
\includegraphics[width=\linewidth]{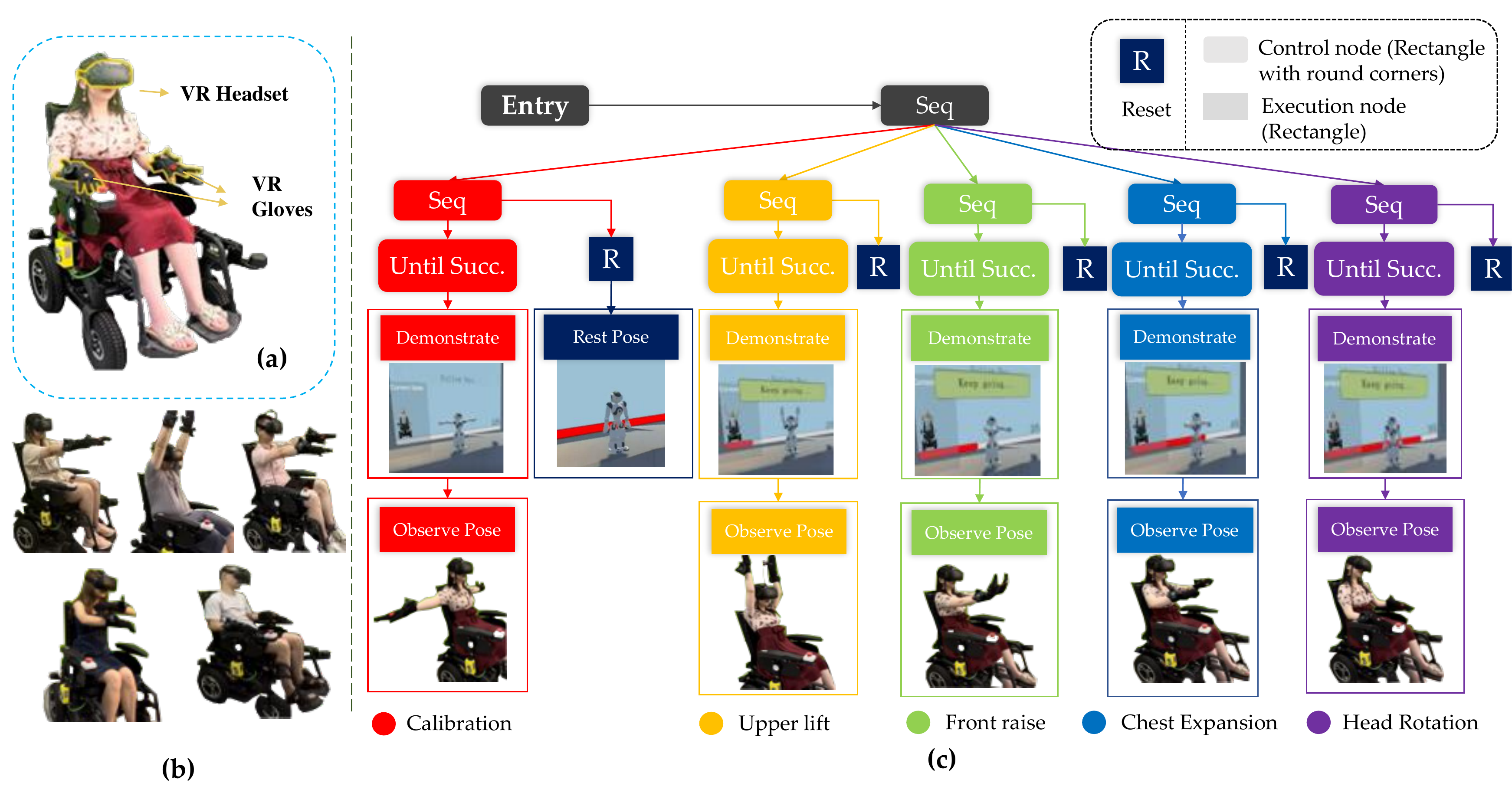}
\end{center}
\vspace{-3mm}
\caption{\footnotesize{Real-world Robotic Social Caregiving: (a) one volunteer with a VR device, (b) other five volunteers, (c) behavior tree for the Nao robot coach.}}
\label{fig:vr}
\vspace{-8mm}
\end{figure}

RCareWorld provides support for social caregiving which is an important direction in robotic caregiving. In this experiment, we program a NAO robot to act as a coach for helping care recipients with daily exercise routines. An exercise consists of a sequence of actions such as lateral raise, upper lift, front raise, chest expansion, and head rotation. We program these exercises in RCareWorld using behavior trees~(Fig.~\ref{fig:vr}). The behavior tree executes the actions of an exercise one-by-one with its sequential nodes. Execution of each action comprises the robot demonstrating the motion, and meanwhile observing the user's action. If the robot decides that the user is following the robot correctly, a green window will pop up to tell the user to keep going. If not, the robot will continue with its demonstration until the user starts following correctly. It decides whether the user is following its demonstration based on the pose of the user. This pose is estimated based on the position of the gloves and headset, and certain predefined constraints on their relative positions. 

During the experiment, a human participant sits in a wheelchair, wears a virtual reality (VR) headset, looks at the NAO coach, and follows its action. With VR gloves, our system records their hand pose, and infers the pose of upper limbs with inverse kinematics.
We show our setting in Fig.~\ref{fig:vr}(b). We ask 6 volunteers to test our system. To test how well our system can recognize users' actions when they follow the robot, we measure the success rate when they perform the actions correctly. The robot can recognize the users' actions with a 100\% success rate. To test how well the system can detect anomalies, we ask the volunteers to do one action wrong in each sequence. It can also recognize the wrong actions with a 100\% success rate. In this user study, we also ask if the users feel comfortable with using VR to provide social caregiving. One user said the VR headset makes her feel dizzy, while others were comfortable with it. All of them could understand and follow the actions of the robot.

%% file: sections/conclusion.tex
\section{Discussion}

In this paper, we propose RCareWorld, a simulation platform that brings
together human avatars and caregiving robots in caregiving home environments to create a human-centric virtual caregiving world. 

RCareWorld uses clinical data to create realistic human models. 
We currently model muscle strength on the MMT data \cite{mmt1}, which is a subjective grading for muscle groups. Access to more detailed numerical measurements such as Electromyography (EMG) data could pave the path towards more realistic models for human avatars. We plan to improve both the physics and speed of human avatar simulations. 
For more realistic physics using soft-tissue modeling, we plan to use Incremental Potential Contact~(IPC)~\cite{Li2020IPC} which provides an intersection and inversion-free modeling for large deformation dynamics. For speed, we plan to implement better support for parallelization on GPU.
 
RCareWorld takes the first step towards a caregiving simulation world with all stakeholders in the loop. With the hope of lowering the barriers to entry to the impactful field of robotic caregiving, we will publicly release the simulation environments with all model assets and human avatars, and keep maintaining it.